\documentclass[10pt,twocolumn,letterpaper]{article}

\usepackage[pagenumbers]{cvpr} 
\definecolor{cvprblue}{rgb}{0.21,0.49,0.74}
\usepackage[pagebackref,breaklinks,colorlinks,allcolors=cvprblue]{hyperref}
\usepackage{cuted}
\usepackage[table]{xcolor}
\usepackage{titletoc}
\usepackage{enumitem}

\usepackage[table]{xcolor} 
\hypersetup{
    colorlinks=true, 
    linkcolor=blue, 
    citecolor=green, 
    urlcolor=magenta 
}

\def\confName{CVPR}
\def\confYear{2026}

\newcommand{\para}{\vspace{0.2cm}}
\newcommand{\bs}{\boldsymbol}

\title{AnyID: Ultra-Fidelity Universal Identity-Preserving Video Generation from \\ Any Visual References}

\author{
    Jiahao Wang\textsuperscript{1}
    \and Hualian Sheng\textsuperscript{2}
    \and Sijia Cai\textsuperscript{2,$\dagger$}%
    \and Yuxiao Yang\textsuperscript{3}
    \and Weizhan Zhang\textsuperscript{1,$*$}%
    \and Caixia Yan\textsuperscript{1}%
    \and Bing Deng\textsuperscript{2}%
    \and Jieping Ye\textsuperscript{2}
}

\date{
    \vspace{0.2cm}
    \textsuperscript{1}School of Computer Science and Technology, MOEKLINNS, Xi’an Jiaotong University\\ \textsuperscript{2}Alibaba Cloud Computing\quad\quad\textsuperscript{3}Tsinghua University\\
    \vspace{0.2cm}
    \texttt{jiahaowang@stu.xjtu.edu.cn}\quad\quad\texttt{stephen.csj@alibaba-inc.com}\quad\quad\texttt {zhangwzh@xjtu.edu.cn}
}

\begin{document}

\maketitle


\begingroup
\renewcommand{\thefootnote}{}
\makeatletter\def\Hy@Warning#1{}\makeatother
\footnotetext{$^*$ Corresponding Author}
\footnotetext{$\dagger$ Project Lead}
\endgroup


\begin{strip}
    \centering
    \vspace{-2.5em}
    \includegraphics[width=\textwidth]{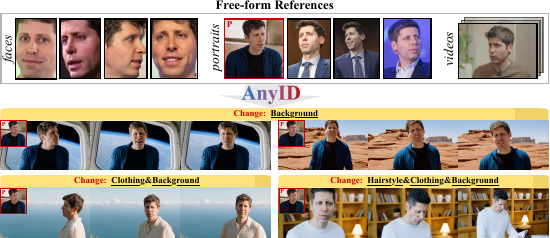}
    \captionof{figure}{AnyID incorporates free-form visual references (e.g., faces, portraits, videos), including one primary reference and several auxiliary references. Controlled by a differential prompt, AnyID is able to perform universal identity-preserving generation based on the primary reference with ultra-high fidelity. The primary reference is boxed in red while the auxiliary references are boxed in black.
    }
    \label{fig:teaser}
\end{strip}

\begin{abstract}
Identity-preserving video generation offers powerful tools for creative expression, allowing users to customize videos featuring their beloved characters. However, prevailing methods are typically designed and optimized for a single identity reference. This underlying assumption restricts creative flexibility by inadequately accommodating diverse real-world input formats. Relying on a single source also constitutes an ill-posed scenario, causing an inherently ambiguous setting that makes it difficult for the model to faithfully reproduce an identity across novel contexts. To address these issues, we present AnyID, an ultra-fidelity identity-preservation video generation framework that features two core contributions. First, we introduce a scalable omni-referenced architecture that effectively unifies heterogeneous identity inputs (e.g., faces, portraits, and videos) into a cohesive representation. Second, we propose a primary-referenced generation paradigm, which designates one reference as a canonical anchor and uses a novel differential prompt to enable precise, attribute-level controllability. We conduct training on a large-scale, meticulously curated dataset to ensure robustness and high fidelity, and then perform a final fine-tuning stage using reinforcement learning. This process leverages a preference dataset constructed from human evaluations, where annotators performed pairwise comparisons of videos based on two key criteria: identity fidelity and prompt controllability. Extensive evaluations validate that AnyID achieves ultra-high identity fidelity as well as superior attribute-level controllability across different task settings. Project page: \url{https://johnneywang.github.io/AnyID-webpage}.
\end{abstract}

\section{Introduction}
The advances of diffusion models \cite{LDM,DDIM,DDPM} have sparked remarkable progress in generative modeling, revolutionizing the creation of visual content, especially in the context of video generation. Large-scale training based on the diffusion transformer (DiT) \cite{DiT} architecture paired with a variational autoencoder (VAE) fosters the generation of visually stunning images and videos, showing potential for a wide spectrum of applications \cite{wang2024oneactor,spotactor,schedule,dynamicid,Animate_Anyone, wan, ControlNet, Phantom, echoshot, yang2025echomotion,yang2025nova3d,yang2025wonder3d,hu2026motionweaver}. Despite the achievements, the realism of universal human-centric video generation (e.g., narrative continuation, short film creation, long movie generation), which features element consistency in identity, hairstyle, clothing, and background across synthesized videos, remains an enticing yet unresolved challenge.

As a pioneering task of human-centric generation, \textit{identity-preserving video generation} \cite{consisid,magicmirror,personalvideo,fantasyid} enables users to customize videos showing their beloved character in various contexts. Along this technical thread, early tuning-based pipelines \cite{dreambooth, magicid, personalvideo} fine-tune the model to learn a specific identity from user-provided images. However, they require laborious tuning every time a new identity is introduced, severely diminishing their usability. Recently, encoder-based pipelines \cite{consisid,magicmirror,Phantom,skyreels} have emerged as a trend for the avoidance of per-identity tuning. They enforce one single facial image as input, introduce a facial encoder leveraging visual expert models \cite{arcface,clip} to inject facial information into the DiT, and conduct training on a curated dataset for generalized identity modeling. However, this single-reference paradigm is fundamentally flawed, which constitutes an ill-posed problem during identity modeling. A person's identity is defined by both their 3D facial structure and their characteristic dynamics of expression. A single 2D snapshot is incapable of capturing this multifaceted information. This inherent ambiguity results in identity deviation when the generated facial angle and expression are different from the reference. For this issue, we argue that the solution lies in embracing multiple free-form references. This approach holds a dual advantage. Technically, it mitigates the ill-posedness by allowing the model to infer 3D shape and dynamic patterns from multiple images or video clips. Practically, it aligns with user behavior, enabling them to leverage their diverse, existing collections of photos and videos to provide a rich, holistic portrait of an individual.

To this end, we present AnyID, a framework for ultra-fidelity universal identity-preserving video generation. Our contributions can be summarized as:
(1) We introduce a novel task formulation that incorporates multiple free-form references. To unify the heterogeneous inputs for consolidated identity modeling, we design a scalable omni-referenced model architecture in which we leverage the original VAE instead of visual experts to inject identity representations for simplicity and broader compatibility. (2) For precise control over attribute variation and invariance, we propose a primary-referenced generation paradigm along with a differential prompt to determine a visual anchor and the expected modification direction. (3) To facilitate supervised training, we construct a human-centric data pipeline based on the PortraitGala dataset \cite{echoshot} to ensure stable and accurate construction of the free-form references and the corresponding differential prompts.
(4) In further pursuit of refined generation quality, we introduce a human-centric reinforcement learning strategy to align the trained model with high-level human expectations. By tailoring two distinct tracks of identity fidelity and prompt controllability, we construct a win-lose dataset to conduct preference tuning for an optimal model performance.

To validate the effectiveness of our framework, we develop a benchmark for a comprehensive evaluation of universal identity-preserving video generation. The results demonstrate that AnyID achieves outstanding dynamic fidelity as well as fine-grained prompt controllability, showing promising potential for diverse applications, which also echoes our initial motivation.

\section{Related Work}
\noindent\textbf{Identity-Preserving Video Generation.} The ability to generate videos of a specific person in diverse contexts is a cornerstone of human-centric applications like virtual avatars and personalized storytelling. Early works often require laborious per-identity fine-tuning \cite{magicid, personalvideo}. To improve usability, recent works have shifted towards an encoder-based paradigm which uses a vision encoder to inject identity features from a single reference image and conducts training on a curated dataset \cite{id-animator,consisid,magicmirror}.
However, this single-reference paradigm is fundamentally limited as it constitutes an ill-posed problem, often resulting in identity drift when generating faces in significantly different angles or with varied expressions. 
In contrast, our work mitigates this limitation by proposing a framework that leverages multiple, free-form references, including images and videos, to build a more complete and robust identity representation.


\para
\noindent{\textbf{Reinforcement Learning in Diffusion Models.}} The success of Reinforcement Learning from Human Feedback (RLHF) in Large Language Models (LLMs) \cite{DPO, PPO, huang2026sketchvl} has inspired numerous efforts to adapt these techniques for visual generation \cite{image-reward, flow-grpo, diffusion-dpo}. Among these, Direct Preference Optimization (DPO) \cite{diffusion-dpo} has emerged as a particularly effective and efficient branch. By optimizing a proxy loss derived directly from preference data, DPO avoids the need for an explicit reward model, significantly reducing training overhead. This approach has been successfully adapted to enhance diffusion models along various dimensions, such as overall visual quality and compositional reasoning.
Building on this, our work introduces a specialized DPO approach for universal identity-preserving video generation, which learns from explicitly disentangled human preferences on identity fidelity and prompt controllability.

\section{Preliminaries}
\textbf{Flow-Based Video Generation.} Prevailing video generation models predominantly employ Rectified Flow (RF) \cite{lipman2022flow,esser2024scaling} to model the distribution of videos. Given a video $\boldsymbol{x}$ along with its description $\boldsymbol{d}$, the video is first mapped to a latent code $\boldsymbol{z}$ through a pre-trained variational autoencoder (VAE) \cite{vae} $\mathcal{E}:\boldsymbol{x}\mapsto \boldsymbol{z}$. In parallel, the description is encoded to a semantic embedding by a pre-trained text encoder $\mathcal{T}:\boldsymbol{d}\mapsto \boldsymbol{c}$. The latent code is then mixed with Gaussian noise $\boldsymbol{\epsilon}$, controlled by a timestep $t$: $\boldsymbol{z}_t=t\boldsymbol{z}_0+(1-t)\boldsymbol{\epsilon}$,
where $\boldsymbol{\epsilon}\sim\mathcal{N}(0,\mathbf{I}), t\in [0,1]$. RF defines the velocity field to progressively connect noise and data, where a parametric network $\boldsymbol{\theta}$, usually a DiT \cite{DiT}, is trained to predict the velocity at a specific timestep $v_t=\frac{dz_t}{dt}=z_0-\epsilon$ given the condition $c$ using the following objective: 
\begin{equation} \label{eq:flow-loss}
\mathcal{L}_\text{RF}(\bs{\theta})=\|v_{\boldsymbol{\theta}}(\boldsymbol{z}_t,t,\bs{c})-v_t\|^2_2.
\end{equation}
Later during inference, Classifier-Free Guidance (CFG) \cite{cfg} combines the conditional and unconditional scores to predict the velocity at each step:
\begin{equation}
    \hat{v}_t=(1-g)\cdot v_{\bs{\theta}}(\bs{z}_t,\bs{t},\bs{c}^\emptyset)+g\cdot v_{\bs{\theta}}(\bs{z}_t,\bs{t},\bs{c}),
\end{equation}
where $g$ is the guidance scale.

\para
\noindent{\textbf{Direct Preference Optimization.}} To incorporate human preferences into generative process, Bradley-Terry (BT) model is adopted to define the preference distribution between a pair of positive and negative generated cases $\{\boldsymbol{x}^+,\boldsymbol{x}^-\}$ under the same condition $\boldsymbol{c}$: $p_{\text{BT}}(\boldsymbol{x}^+\succ\boldsymbol{x}^-|\boldsymbol{c})=\sigma(r(\boldsymbol{x}^+,\boldsymbol{c})-r(\boldsymbol{x}^-,\boldsymbol{c}))$, where $\sigma(\cdot)$ is the sigmoid function and the reward $r(\boldsymbol{c},\boldsymbol{x})$ measures the utility of each case. In the context of latent flow models, Diffusion-DPO \cite{diffusion-dpo} presents a tractable formulation to optimize the reward via implicit reward modeling:
\begin{equation} \label{eq:dpo}
    \mathcal{L}_\text{Flow-DPO}(\bs{\theta})=\log\sigma(-\beta(s(\bs{z^+_t,t,\bs{c},\bs{\theta}}))-s(\bs{z^-_t,t,\bs{c},\bs{\theta}}))),
\end{equation}
\begin{equation}
    s(\bs{z}_t,t,\bs{c},\bs{\theta})=\|v_{\bs{\theta}}(\bs{z}_t,t,\bs{c})-v_t\|^2_2-\|v_{\text{ref}}(\bs{z}_t,t,c)-v_t\|^2_2,
\end{equation}
where $\beta$ is a hyperparameter that controls regularization.

\section{Method}
\begin{figure*}[t]
\centering
\includegraphics[width=\linewidth]{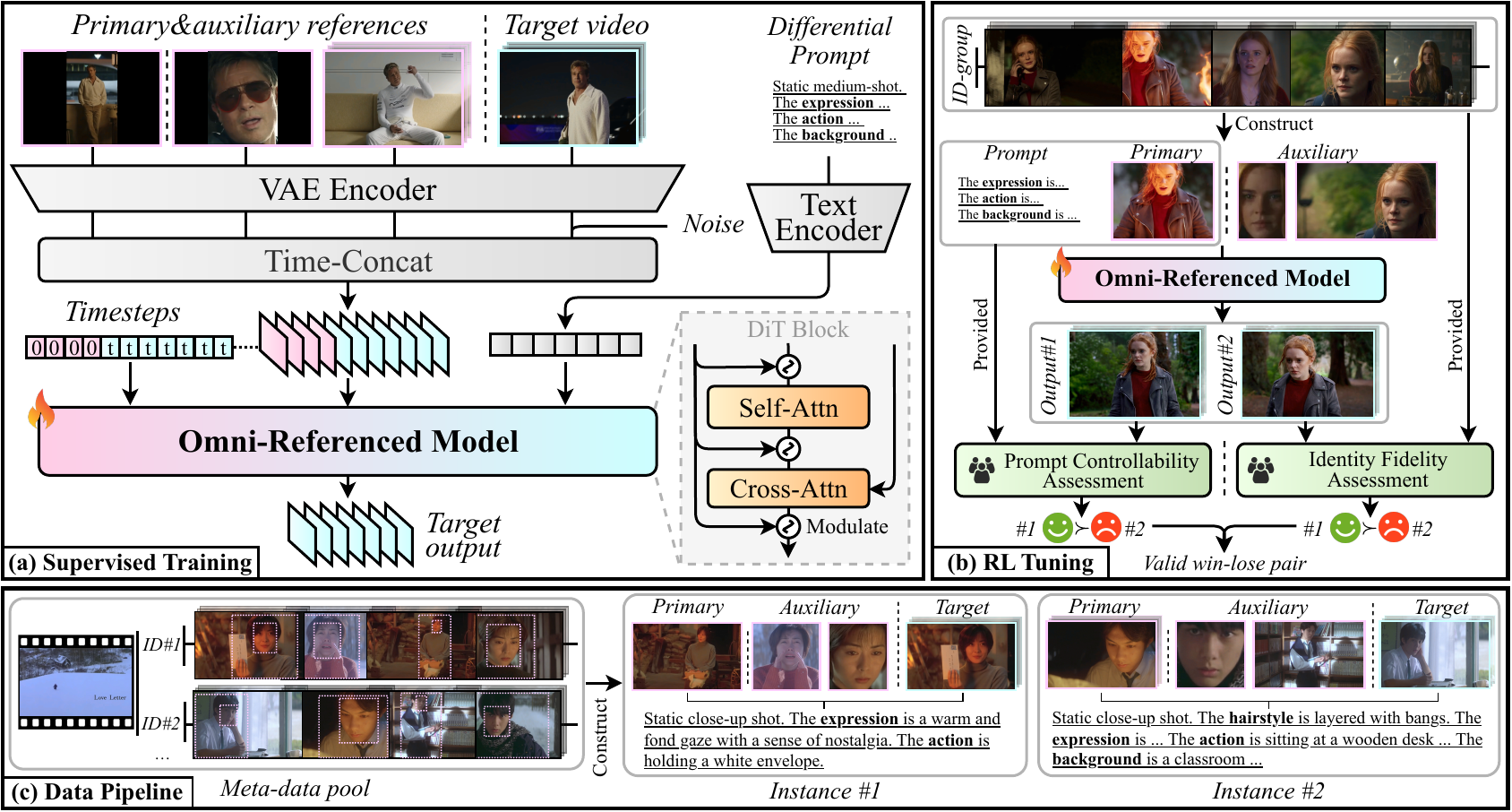}
\caption{The overall framework of AnyID consists of two stages of training: (a) During supervised tuning, we propose a scalable omni-referenced model to unify the identity inputs and design a primary-referenced generation to enable precise control. (b) RL tuning is further performed to enhance the identity fidelity and prompt controllability. (c) The training is facilitated by a meticulously devised data pipeline.}
\label{fig:pipeline}
\end{figure*}

To achieve universal identity-preserving generation, we begin by harmonizing various tasks and expanding their boundaries with a free-form reference formulation in Sec.~\ref{sec:task}. We then design a scalable omni-referenced framework to unify heterogeneous input forms in Sec.~\ref{sec:model}. For precise prompt controllability, we propose a primary-referenced generation paradigm along with the concept of differential prompt in Sec.~\ref{sec:primary}. In pursuit of optimal identity fidelity and prompt controllability, we further perform a human-centric reinforcement learning in Sec.~\ref{sec:dpo}. The supervised training is facilitated by a multi-reference data curation pipeline in Sec.~\ref{sec:data}.

\subsection{Task Formulation} \label{sec:task}
Single-reference video generation is an inherently ill-posed problem, where the ambiguity of inferring 3D structure and dynamics from a static image leads to severe artifacts and low fidelity. Therefore, in this paper, the objective of universal identity-preserving video generation is to synthesize a video $\bs{x}$ that preserves a person's identity from a set of reference inputs while adhering to a given prompt. Formally, this problem has two sets of inputs:
\begin{itemize}
    \item A Reference Set ${\mathcal{R}}$: A collection of $N$ visual inputs, $\mathcal{R}=\{\bs r_i\}_{i=1}^N$, that provide information about a person's identity and appearance. These references can be of mixed modalities, such as static images and video clips, capturing the person in various contexts.
    \item {A Target Prompt $\bs{d}$: A textual description specifying the target video's content and dynamics.}
\end{itemize}

Given these inputs, the objective is to generate a target video $\bs{x}$ that simultaneously satisfies two core constraints: identity preservation and prompt controllability. 

\subsection{Omni-Referenced Identity-Preserving Model} \label{sec:model}
\textbf{Unifying Diverse Identity References.} Given a set of $N$ free-form visual references $\mathcal{R}$, the central challenge is how to harmonize the heterogeneous conditions for consolidated identity modeling. Existing methods often rely on dedicated external encoders (e.g., face or CLIP encoders) to embed identity information. However, this approach often creates an architectural bottleneck, which limits generative fidelity and inherently struggles with diverse input formats.
In contrast, we propose a simple yet scalable \textit{omni-referenced approach} that leverages the generalization capability of a single, pretrained VAE encoder. Our process begins with input standardization, as illustrated in Fig.~\ref{fig:pipeline}(a). All visual references, with static images treated as single-frame videos, are first padded to a uniform resolution of $H\times W$ while preserving their original aspect ratios. Subsequently, each standardized input $\bs{r}_i$ is independently encoded into latent representations using the VAE encoder $\mathcal{E}(\cdot)$. These latent codes are then concatenated along the temporal dimension to form a unified condition:
\begin{align}
    \bs{y}=\mathrm{Concat}(\{\mathcal{E}(\mathcal{I}(\bs{r}_i))\}_{i=1}^N),
\end{align}
where $\mathcal{I}(\cdot)$ denotes the resizing and padding operation. Afterwards, given the corresponding ground-truth video-description pair $\{\bs{x},\bs{d}\}$ during training, we encode the target video into $\bs{z}_0=\mathcal{E}(\bs{x})$ and mix it with Gaussian noise $\bs{\epsilon}$ to obtain $\bs{z}_t$. The description $\bs{d}$ is also encoded into $\bs{c}$ by a text encoder. During the forward process, the clean reference latents and the noisy target latent are also concatenated along the temporal dimension. To enable the model to distinguish between these two inputs, we employ a timestep manipulation strategy. Specifically, we assign a static timestep of $t=0$ to all the reference latents, indicating they are noise-free. In contrast, the target latent is assigned its corresponding sampled timestep $t>0$. When the composite sequence, along with the description $\bs{c}$ and timestep embeddings, is fed into the DiT backbone, the model learns to process them differently. The network's output components corresponding to the zero-timestep references are discarded, while the output for the target is used to compute the final RF loss:
\begin{equation}
\mathcal{L}_\text{RF}(\bs{\theta})=\|v_{\boldsymbol{\theta}}(\bs{z}_t,\bs{y},\bs{t},\bs{c})-(\bs{z}_0-\bs{\epsilon})\|^2_2,
\end{equation}

\para
\noindent \textbf{Aggregating Unconditional Guidance.} Prior works like \cite{Phantom} often handle multiple conditions by performing a separate inference pass for each, which significantly increases computational overhead. To overcome this limitation, we establish a single, unified unconditional state for guidance by simultaneously nullifying all modalities. Specifically, this is achieved by inputting the latent code of pure black pixels $\bs{y}^\emptyset$ visually and the embedding of an empty prompt $\bs{c}^\emptyset$ semantically.
The predicted velocity during inference is:
\begin{equation}
    \hat{v}_t=(1-g)\cdot v_{\bs{\theta}}(\bs{z}_t,\bs{y}^\emptyset,\bs{t},\bs{c}^\emptyset)+g\cdot v_{\bs{\theta}}(\bs{z}_t,\bs{y},\bs{t},\bs{c}),
\end{equation}
where $g$ is the guidance scale. This design avoids extra inference computation and functions well in practice.

\subsection{Primary-Referenced Generation} \label{sec:primary}
\noindent\textbf{Generating Differential Prompt.} Establishing a person's identity from multiple visual references introduces an inherent challenge: how to reconcile contextual inconsistencies across the references. Variations in lighting, makeup, and background can indicate conflicting representations of a person's attributes (e.g., makeup, hairstyle, clothes). To resolve this conflict, we propose a \textit{primary-referenced generation} paradigm. In this framework, we designate a single source, the primary reference ($\bs{r}_1$), as the canonical anchor for all static attributes. This novel primary-referenced structure necessitates a new way of specifying transformations. Conventional prompting strategies, which rely on absolute, overall descriptions of the target scene, are ill-suited for this task. Such prompts are not only laborious for users to craft but also struggle to precisely describe one specific change relative to the primary reference, often leading to unexpected alterations. To address this issue, we introduce the concept of a \textit{differential prompt}. 
To specify, the differential prompt describes only the changes from the primary reference to the target content, while anything not mentioned in it is expected to remain consistent.
This differential strategy minimizes descriptive noise by focusing the model's attention exclusively on specified changes while ensuring a stable contextual foundation by anchoring every transformation to a single reference point.

During practice, we dissect the video content into seven predefined \textit{portrait-video elements}: shot type, hairstyle, clothes, accessory, expression, action, and background. Our differential prompt generation process unfolds in two stages. First, for a given primary reference $\bs{r}_i$ and the target video $\bs x$, we employ a VLM to generate detailed textual descriptions for each element, resulting in a reference description set $\bs{d}_\textit{ref}$ and a target description set $\bs{d}_\textit{tar}$. Second, we use an LLM to perform a comparative analysis. The LLM quantifies the semantic similarity for each corresponding element pair between $\bs{d}_\textit{ref}$ and $\bs{d}_\textit{tar}$. Elements with a similarity score exceeding a threshold $\gamma$ are deemed unchanged and are pruned from $\bs{d}_\textit{tar}$. The remaining descriptions constitute the final differential prompt $\bs d$. Our experiments show that directly using a single VLM for differential prompting is currently unstable, often leading to hallucinations and omissions. Therefore, our two-stage pipeline offers a more robust solution by decomposing the task. Even so, our core contribution is introducing the concept of differential prompting, not its specific extraction method. A sufficiently advanced VLM could potentially perform this task directly in the future.

\subsection{Human-Centric Reinforcement Learning} \label{sec:dpo}


While the supervised training in existing works with a pixel-wise  Mean Squared Error (MSE) loss establishes a strong baseline, it is fundamentally misaligned with high-level human perception. MSE often leads to overly smooth outputs that degrade the fine-grained features essential for both identity fidelity and prompt controllability. To overcome this limitation, we are in need of a learning strategy that optimizes directly based on human perceptual judgment. Reinforcement learning (RL), particularly when guided by human feedback, serves as an ideal solution to bridge this gap between pixel-level metrics and user-centric goals. To this end, we introduce a human-centric reinforcement learning framework, shown in Fig.~\ref{fig:pipeline}(b), which uses human preference annotations as its optimization objective. 

Concretely, we first extract numerous sets of inputs from the training dataset and sample two videos for each set using our trained model to form a matchup. Then, paired samples are evaluated along two tracks: omni-referenced identity fidelity and primary-referenced prompt controllability. To assess identity fidelity, annotators are provided with the complete set of clips from the source ID-group as a comprehensive reference. They are then tasked with selecting the video that better preserves the person's dynamic identity. For the primary-referenced prompt controllability evaluation, annotators review the primary reference and the differential prompt. Their task is to select the video that better aligns with the prompt's intended transformation while preserving consistency in aspects not specified by the prompt. A pair is deemed a valid training sample only if one video is unequivocally superior to the other across both evaluation axes (i.e., achieves Pareto dominance). We collect 1,000 valid pairs to fine-tune our model via reinforcement learning with the DPO objective in Eq.~(\ref{eq:dpo}).

\subsection{Multi-Reference Data Curation} \label{sec:data}
\textbf{Building Meta-Data Pool.} 
The foundation for achieving ultra-high fidelity is a dataset that enables the model to learn identity invariance across diverse and complex scenarios. 
To this end, we construct a large-scale human identity meta-data pool by curating the PortraitGala dataset \cite{echoshot}. While PortraitGala inherently groups clips by person identity (ID-groups), we apply a rigorous filtering pipeline to enhance data quality. This process involves two steps: first, we apply face detection to isolate single-subject clips, and second, we filter out low-quality samples, such as those with blurry faces or extreme head poses. To further enrich the dataset, the annotations of each video are augmented with its corresponding face and person detection outputs (e.g., bounding boxes). This curation process yields a meta set of 100k ID-groups and 300k video clips. 



\para
\noindent\textbf{Constructing Diverse Reference Forms.} While the curated data pool provides high-quality, identity-consistent video clips, its homogeneity does not reflect the diversity of real-world inputs. To bridge this gap, we implement a data augmentation strategy to synthesize these varied reference forms from our curated clips.

Starting from one ID-group of clips, we first sample $N$ reference videos and a target video $\bs{x}$. To simulate real-world scenarios with varied reference formats, we transform each of the $N$ reference videos into one of three forms (i.e., face, portrait, or video) as follows:
\begin{itemize}
    \item Face image reference: The facial region is cropped from a single, randomly selected frame.
    \item Portrait image reference: A wider or tighter portrait is generated by cropping the region around the person from a random frame.
    \item Video reference: A random video segment is sampled, and its unified person region is subsequently cropped.
\end{itemize}
In summary, each iteration of this data construction pipeline yields a complete training instance: a primary reference, a set of auxiliary references in a variety of formats, and a target video. This multi-faceted data structure is crucial for endowing our model with the versatility to handle heterogeneous reference inputs.




\begin{figure*}[t]
\setlength{\abovecaptionskip}{0.2cm}
\setlength{\belowcaptionskip}{-0.1cm}
\centering
\includegraphics[width=1\linewidth]{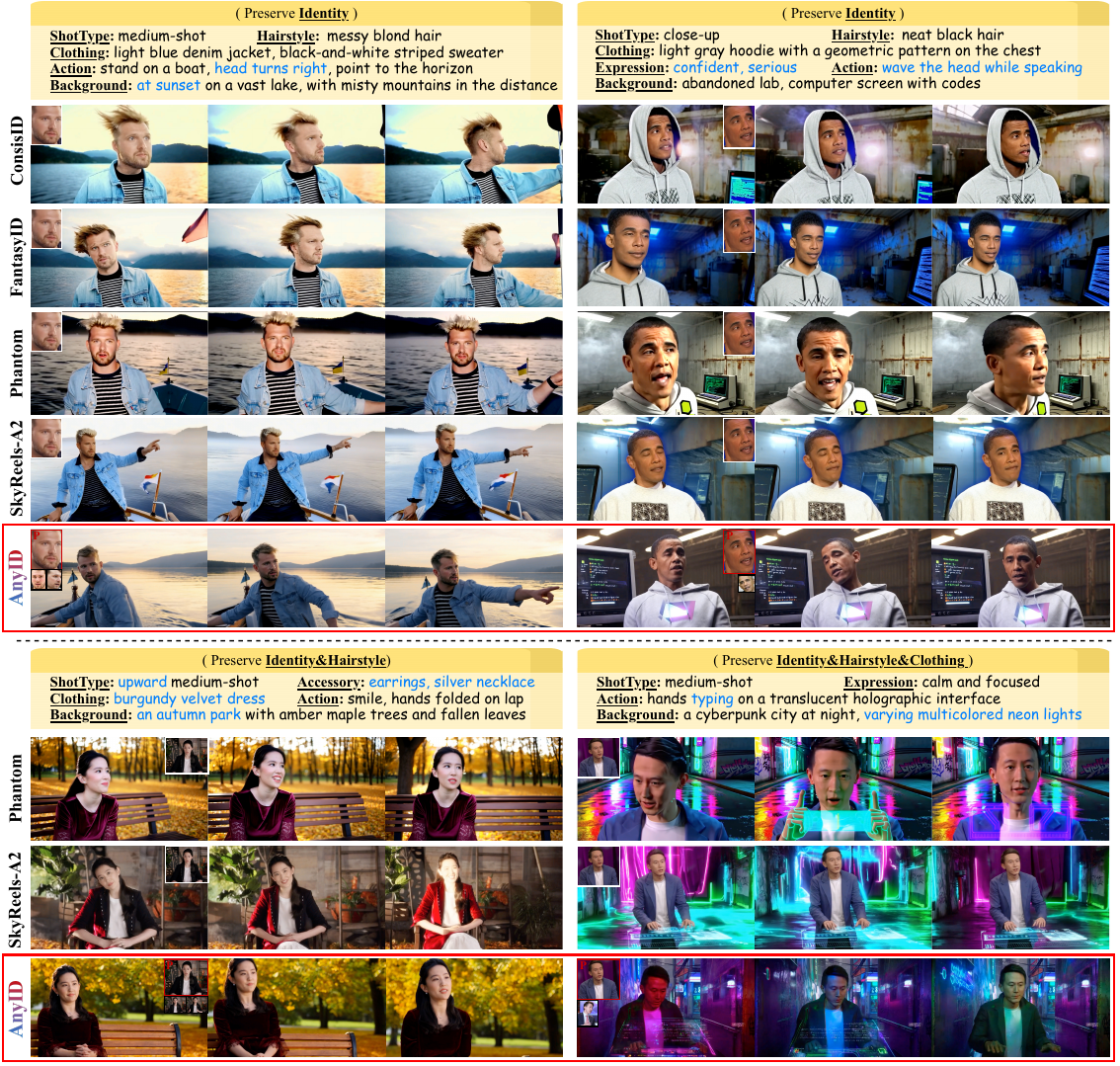}
\caption{Illustration of AnyID and baselines. Across different settings, AnyID continuously demonstrates ultra-high identity fidelity in varying facial angles and motions as well as superior prompt controllability, benefiting from the free-form references. The primary reference is boxed in red while the auxiliary references are boxed in black. Please zoom in for a better view.}
\label{fig:main}
\end{figure*}

\section{Experiments}
\textbf{Implementation Details.} AnyID is implemented based on Wan-5B \cite{wan}. We set the resolution to 1280$\times$704 and the frame number to 121, which is equivalent to 5 seconds of 720p video. During training and inference, the number of references $N$ varies from 1 to 5. The probability of null conditions $p_\emptyset$ is set to 0.1, and the guidance scale $g$ is set to 5.0. The supervised training and RL tuning are both conducted in the LoRA \cite{lora} manner. The training of AnyID takes about 4,500 NVIDIA A100 GPU hours in total.

\para
\noindent\textbf{Baselines.} 
To comprehensively evaluate the performance of AnyID, we compare it against a diverse set of baselines from both academia and industry:
(1) ConsisID \cite{consisid} and FantasyID \cite{fantasyid}, both implemented based on CogvideoX-5B \cite{cogvideo}; (2) Phantom \cite{Phantom} and SkyReels-A2 \cite{skyreels}, both implemented based on Wan-14B \cite{wan}.
The configurations for all the baselines remain set to default.

\begin{table*}[t]
\setlength{\abovecaptionskip}{0.1cm}
\setlength{\belowcaptionskip}{-0.1cm}
\caption{Metric results of AnyID, baselines, and ablation models. The best and second-best scores are denoted \textbf{bold} and \underline{underlined}.}
\label{tab}
\centering
\setlength{\tabcolsep}{2.5mm}{
\begin{tabular}{l|cc|cc|ccc|cc}
\hline
                                            & \multicolumn{2}{c|}{\textit{Identity fidelity}}             & \multicolumn{2}{c|}{\textit{Element consistency}}           & \multicolumn{3}{c|}{\textit{Prompt controllability}}                                       & \multicolumn{2}{c}{\textit{Visual quality}}                 \\ \cline{2-10} 
                                            & Holi-Arc                  & Holi-Cur                  & Ele-CLIP                         & Ele-DINO                         & App.                         & Mot.                         & Bg.                          & Sta.                         & Dyn.                         \\ \hline
ConsisID-5B                                 & 68.92                        & {\underline{65.07}}                  & -                            & -                            & 66.56                        & 51.88                        & 78.54                        & 74.75                        & 75.37                        \\
FantasyID-5B                                & 69.60                        & 65.03                        & -                            & -                            & {\underline{73.12}}                  & 58.74                        & 82.50                        & 73.81                        & 82.81                        \\ \hline
Phantom-14B                                 & {\underline{70.01}}                  & 64.01                        & 66.01                        & {\underline{75.90}}                  & 61.88                        & 54.37                        & {\underline{83.12}}                  & 77.62                        & {\underline{83.54}}                  \\
SkyReels-A2-14B                             & 69.12                        & 64.89                        & \textbf{78.14}               & \textbf{79.87}               & 50.19                        & {\underline{58.75}}                  & 54.37                        & {\underline{78.62}}                  & 78.25                        \\ \hline
{\color[HTML]{9B9B9B} AnyID-5B w/o P\&D} & {\color[HTML]{9B9B9B} 72.58} & {\color[HTML]{9B9B9B} 66.90} & {\color[HTML]{9B9B9B} 63.87} & {\color[HTML]{9B9B9B} 64.03} & {\color[HTML]{9B9B9B} 80.82} & {\color[HTML]{9B9B9B} 56.45} & {\color[HTML]{9B9B9B} 83.60} & {\color[HTML]{9B9B9B} 81.81} & {\color[HTML]{9B9B9B} 89.39} \\
{\color[HTML]{9B9B9B} AnyID-5B w/o RL}     & {\color[HTML]{9B9B9B} 72.61} & {\color[HTML]{9B9B9B} 66.95} & {\color[HTML]{9B9B9B} 65.49} & {\color[HTML]{9B9B9B} 65.96} & {\color[HTML]{9B9B9B} 80.77} & {\color[HTML]{9B9B9B} 57.80} & {\color[HTML]{9B9B9B} 83.92} & {\color[HTML]{9B9B9B} 81.79} & {\color[HTML]{9B9B9B} 90.44} \\
{\color[HTML]{9B9B9B} AnyID-5B w/o AR}     & {\color[HTML]{9B9B9B} 72.38} & {\color[HTML]{9B9B9B} 66.77} & {\color[HTML]{9B9B9B} 68.61} & {\color[HTML]{9B9B9B} 69.55} & {\color[HTML]{9B9B9B} 86.19} & {\color[HTML]{9B9B9B} 60.30} & {\color[HTML]{9B9B9B} 83.24} & {\color[HTML]{9B9B9B} 81.98} & {\color[HTML]{9B9B9B} 91.11} \\
AnyID-5B (ours)                             & \textbf{73.22}               & \textbf{67.52}               & {\underline{68.78}}                  & 69.50                        & \textbf{86.56}               & \textbf{60.31}               & \textbf{84.79}               & \textbf{82.03}               & \textbf{91.12}               \\ \hline
\end{tabular}
}
\end{table*}

\subsection{Benchmark}
\para
\noindent\textbf{Evaluation Set.} To facilitate a comprehensive and rigorous performance evaluation,
we propose a benchmark tailored
for universal identity-preserving video generation. To start with, we define two types of tasks: \textit{identity-preserving generation} (IPT2V), which only preserves identity, and \textit{identity-element-preserving generation} (IEPT2V), which preserves identity along with certain elements. We then collect 50 free-form reference sets of 50 celebrities, each containing 5 visual references showing various facial angles and motions. We instruct LLM to accordingly generate 50 IPT2V prompts and 50 IEPT2V prompts, preserving hairstyle, hairstyle\&clothing, or hairstyle\&clothing\&background. Note that since the baselines are incompatible with multiple references, we only input the primary reference. ConsisID and FantasyID don't participate in the IEPT2V task. 

\para
\noindent\textbf{Metrics.}
We further assess quantitative metrics across four dimensions: (1) \textit{Identity fidelity}: 
Different from conventional single-reference benchmarks, we calculate the average normalized cosine similarity between the generated face and all the reference faces utilizing ArcFace \cite{arcface} and CurricularFace \cite{curricularface} to report more trustworthy holistic metrics, Holi-Arc and Holi-Cur. Such a measurement reflects the overall fidelity of the generated character across different facial angles. (2) \textit{Element consistency}: We utilize Grounded-SAM \cite{grounded-sam} to segment all the elements expected to be preserved from the primary reference and the generated video and then compute the normalized cosine similarity between their embeddings utilizing CLIP \cite{clip} and DINO \cite{dino} to report Ele-CLIP and Ele-DINO. (3) \textit{Prompt controllability}: To reflect motion conformity and fine-grained controllability, unlike conventional benchmarks that measure the visual-semantic similarity, we turn to VLM to score the alignment between the prompt and the whole video in three aspects: human appearance (App.), camera and human motions (Mot.), background (Bg.). (4) \textit{Visual quality}: We follow Vbench \cite{vbench} to instruct VLM to score in two aspects. Static score (Sta.) covers clarity, color saturation, content layout, etc. Dynamic score (Dyn.) covers smoothness, temporal consistency, reasonableness, etc.

\begin{figure*}[t]
\centering
\includegraphics[width=1\linewidth]{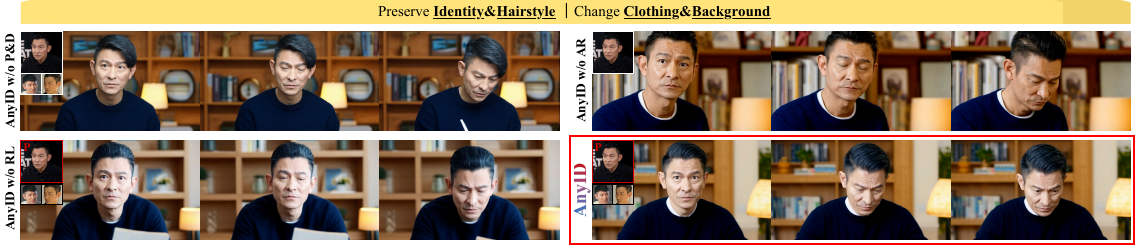}
\setlength{\abovecaptionskip}{-0.2cm}
\setlength{\belowcaptionskip}{-0.3cm}
\caption{Generated cases by AnyID and three ablation models, which confirms that our proposed components function as expected}
\label{fig:ablation}
\end{figure*}
\begin{figure}[t]
\setlength{\abovecaptionskip}{0.2cm}
\setlength{\belowcaptionskip}{-0.4cm}
\centering
\includegraphics[width=\linewidth]{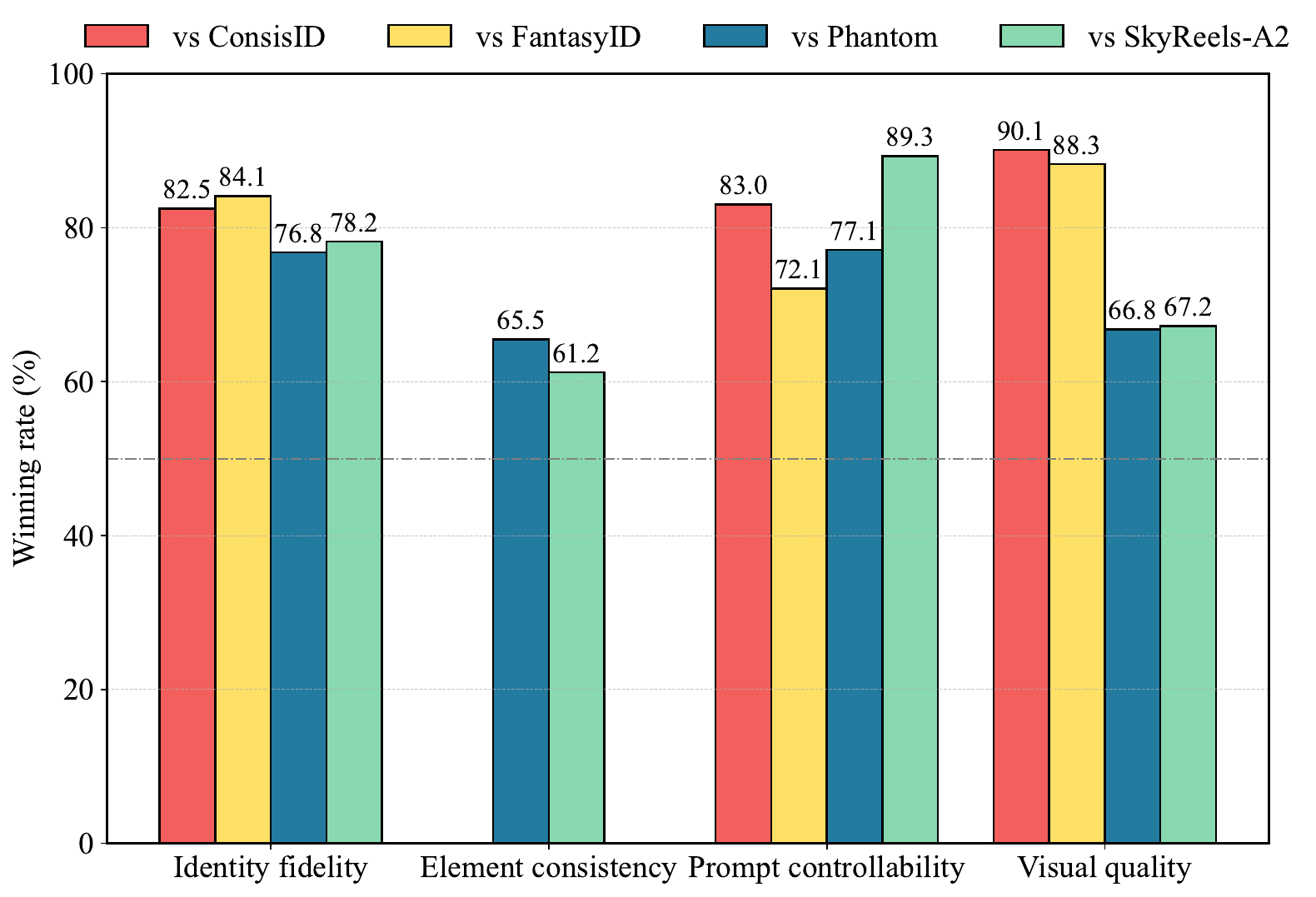}
\caption{Results of user study. The overall high winning rate proves that AnyID better aligns with human preferences.}
\label{fig:userstudy}
\end{figure}
\subsection{Qualitative Evaluation} \label{sec:quali}
\textbf{IPT2V.} In the upper part of Fig.~\ref{fig:main}, we illustrate the qualitative comparison in IPT2V task. With only one facial reference, the negative effects of ill-posedness are evident across all baselines: (1) Inferior dynamic fidelity is observed in ConsisID, FantasyID, and Phantom with significant identity deviation when the head rotates or when the facial expression changes differently from the reference image. (2) A severe bias toward the reference image is observed in SkyReels-A2, ignoring the prompt control and even causing visual distortion. In contrast, our AnyID demonstrates superior identity fidelity across varying facial angles and motions. This benefits from the rich multi-view information provided by free-form references, which resonates with our initial motivation.

\para
\noindent\textbf{IEPT2V.} The lower part of Fig.~\ref{fig:main} illustrates the performance in IEPT2V task. As shown, Phantom continuously exhibits identity distortion across varying facial angles and expressions. While for SkyReels-A2, the bias issue still persists with unexpected "copy-paste" of clothes and even the background. By comparison, in addition to showcasing ultra-high facial fidelity as always, our AnyID simultaneously achieves the optimal balance of element consistency and prompt controllability. For the example in the bottom-right corner of Fig.~\ref{fig:main}, AnyID not only preserves the identity, hairstyle, and clothing but also renders the most natural visual effect according to the described environment and lighting, showcasing unparalleled quality. This proves that AnyID effectively models the identity invariance across varying contexts and aligns with human preferences, owing to the supervised training and RL tuning.

\subsection{Quantitative Evaluation}
\textbf{Metric Results.} We report the quantitative results of AnyID and baselines in Tab.~\ref{tab}. Our method scores the highest 73.22\% on Holi-Arc and 67.52 \% on Holi-Arc, demonstrating ultra-high dynamic identity fidelity. For subject consistency, the top performance of SkyReels-A2 is a result of the "copy-paste" phenomenon, as discussed in Sec.~\ref{sec:quali}. For Phantom, we observe a divergence in scores between Ele-CLIP and Ele-DINO. We speculate that this is because Phantom's visual results exhibit a stylistic tendency, altering the high-frequency information of elements, which is captured by CLIP but not by DINO. In terms of prompt controllability, our method tops the scoreboard with a remarkable 86.56\% App., 60.31\% Mot., and 84.79\% Bg. Finally, our method leads with 82.03\% Sta. and 91.12\% Dyn., showing superior overall visual quality.

\para
\noindent\textbf{User Study.}
To assess the performance from the perspective of human preference, we further conduct a user study involving 20 participants. We instruct each of them to perform binary voting in 40 one-on-one matchups between AnyID and the baselines across the above-mentioned four dimensions. As shown in Fig.\ref{fig:userstudy}, AnyID wins more than 70\% matchups in identity fidelity and prompt controllability and wins more than 60\% matchups in element consistency and visual quality, showing an absolute advantage. Among the results, the victory in element consistency comes as a surprise. During the post interviews, it was revealed that many participants included naturalness in their subjective judgment of element consistency, leading to a preference toward our method. In general, the user study reconfirms the excellence of AnyID, which echoes the qualitative results.

\subsection{Ablation Study}
We conduct an ablation study to verify the effectiveness of the proposed architecture. To specify, we construct three ablation models: (1) our method excluding the primary-reference and differential-prompt design, using the ordinary prompt (AnyID w/o P\&D);
(2) our method excluding reinforcement learning (AnyID w/o RL); (3) our method excluding auxiliary references during inference (AnyID w/o AR). We report the quantitative ablation results in Tab.~\ref{tab}. Observing the case shown in Fig.~\ref{fig:ablation}, we can draw three conclusions: (1) P\&D enables the model to resolve the hairstyle conflicts across multiple references, faithfully preserving the expected hairstyle. (2) AR provides additional facial dynamic information during inference, facilitating precise modeling of identity invariance across diverse facial motions. (3) RL refines the performance in aspects that subtly affect fidelity, such as hair shine, texture, and facial shading.
These three conclusions confirm that our proposed architecture is practically effective and aligns with expectations.

\section{Conclusion}
We present AnyID, an identity-preserving video generation framework that addresses the key limitations of existing methods. We identify that the reliance on one single image reference causes identity ambiguity and limits creative flexibility. For these issues, we design a scalable omni-referenced architecture that unifies heterogeneous inputs and a primary-referenced generation paradigm that uses differential prompting for anchored control. To further refine the performance, we employ a reinforcement learning stage based on human feedback. Comprehensive experiments demonstrate that our method significantly outperforms existing baselines, particularly in the critical dimensions of identity fidelity and prompt controllability.

\nocite{li2026planviz,huang2026sketchvl}

{
    \small
    \bibliographystyle{ieeenat_fullname}
    \bibliography{main}
}


\clearpage

\onecolumn
\thispagestyle{empty}
\appendix

\begin{center}
    \LARGE\textbf{Appendix}

    \vspace{0.3cm}
\end{center}

\startcontents[sections]
\section*{Table of Contents}
\vspace{0pt}\noindent\hrulefill

\printcontents[sections]{l}{1}{\setcounter{tocdepth}{2}}

\noindent\hrulefill

\newpage

\setcounter{page}{13}
\twocolumn

\section{Details of Dataset Construction}
\begin{figure}[h]
\centering
\includegraphics[width=1\linewidth]{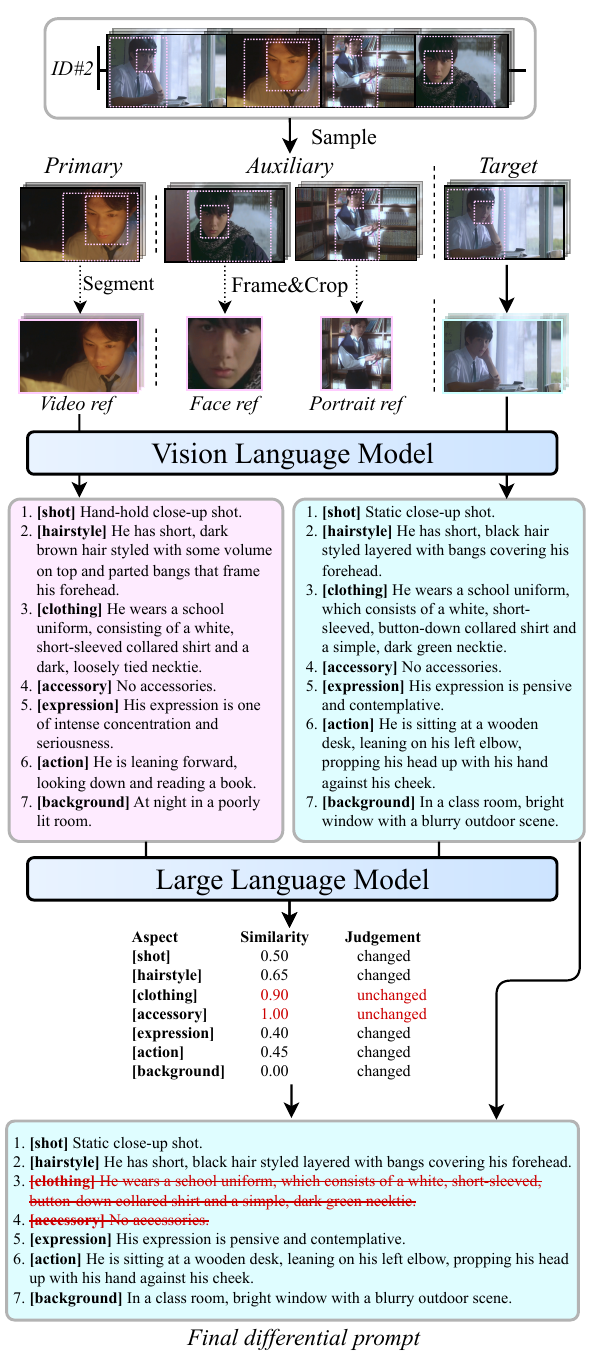}
\caption{A construction example of the visual references and the differential prompt for the supervised training.}
\label{app:fig:datacase}
\end{figure}

\newpage

In our paper, we introduce a data pipeline for constructing diverse visual references and differential prompts to facilitate supervised training. Here, we provide a detailed description with a construction example in Fig.~\ref{app:fig:datacase}. As illustrated, the whole pipeline starts from sampling one video as the primary reference, several videos as auxiliary references, and one video as the target. All the reference videos randomly undergo operations to produce different forms of visual references. Temporal segmentation produces video references. Sampling one frame and cropping the facial region produces face image references. Sampling one frame and cropping the region around the person produces portrait image references. Afterwards, the primary reference and the target video are respectively input to a VLM for detailed descriptions in the seven portrait-video aspects. The two descriptions are then input to an LLM to quantify the similarity between them in terms of the seven aspects. Aspects with a similarity score exceeding a threshold $\gamma$, which is set to 0.8 in practice, are deemed unchanged. The unchanged aspects are removed from the description of the target video to produce the final differential prompt.

\section{More Experimental Details}
We utilize the text input mode of Gemini-2.5-Pro as the LLM and the text-video input mode as the VLM. All the training is carried out on NVIDIA A100 80GB GPUs. The supervised training starts from standard Wan-5B using LoRA, which takes 4,000 GPU hours. The RL tuning starts from the merged weights after supervised training, also using LoRA, which takes additional 500 GPU hours. Throughout the training, all the important settings are listed below:

\begin{table}[h]
\centering
\rowcolors{2}{gray!20}{white} 
\caption{Experimental settings of AnyID.}
\setlength{\tabcolsep}{10mm}{
\begin{tabular}{cc}
\hline
\textbf{Parameter}    & \textbf{Value} \\ \hline
Video height          & 704            \\
Video width           & 1280           \\
Video frame           & 121            \\
FPS                   & 24             \\
LoRA rank             & 128            \\
LoRA alpha            & 128            \\
Batchsize             & 64             \\
Train timesteps       & 1000           \\
Train shift           & 5.0            \\
Optimizer             & AdamW          \\
Learning rate         & 1e-4           \\
Weight decay          & 0.001          \\
Sample timesteps      & 50             \\
Sample shift          & 5.0            \\
Sample guidance scale & 5.0            \\ \hline
\end{tabular}
}
\label{tab:settings}
\end{table}

\section{Details of Benchmark}
 \textbf{Celebrities.} To construct a comprehensive and objective evaluation set, we must initially choose target individuals with consistent appearances, abundant and easily accessible visual records, and preferably well-known to the public, for better intuitive evaluation of the generated fidelity. Hence, we utilized LLM to generate a list of 50 celebrities with a balanced distribution of ethnicity and gender, spanning diverse fields such as AI academia, entrepreneurship, athletics, acting, and music. 
 The list of celebrities is as follows:

\begin{itemize}[leftmargin=*,labelindent=20pt] 
    \item 001 Sam Altman  
    \item 002 Jensen Huang  
    \item 003 Elon Musk  
    \item 004 Jun Lei  
    \item 005 Yun Ma  
    \item 006 Shou Zi Chew  
    \item 007 Feifei Li  
    \item 008 Yann LeCun  
    \item 009 Stephen Curry  
    \item 010 Ming Yao  
    \item 011 Kobe Bryant  
    \item 012 Luka Doncic  
    \item 013 LeBron James  
    \item 014 Lionel Messi  
    \item 015 Cristiano Ronaldo  
    \item 016 Long Ma  
    \item 017 Hanyu Yuzuru  
    \item 018 Eileen Gu  
    \item 019 Ziyi Zhang  
    \item 020 Yuyan Peng  
    \item 021 Andy Lau  
    \item 022 Yi Zhang  
    \item 023 Liying Zhao  
    \item 024 Jackie Chan  
    \item 025 Crystal Liu  
    \item 026 Benedict Cumberbatch  
    \item 027 Morgan Freeman  
    \item 028 Leonardo DiCaprio  
    \item 029 Robert Downey Jr.  
    \item 030 Will Smith  
    \item 031 Emma Watson  
    \item 032 Andrew Garfield  
    \item 033 Rowan Atkinson  
    \item 034 Anne Hathaway  
    \item 035 Scarlett Johansson  
    \item 036 Keira Knightley  
    \item 037 Halle Berry  
    \item 038 Satomi Ishihara  
    \item 039 Hiroshi Abe  
    \item 040 Jay Chou  
    \item 041 Rihanna  
    \item 042 Taylor Swift  
    \item 043 Eason Chan  
    \item 044 Billie Eilish  
    \item 045 JJ Lin  
    \item 046 Mika Nakashima  
    \item 047 Lana Del Rey  
    \item 048 Ed Sheeran  
    \item 049 Bruno Mars  
    \item 050 Beyoncé  
\end{itemize}

\para
\noindent\textbf{Visual References.} For each celebrity, we elaborately collect four portrait images and one video from the website, ensuring that they are free of visual filters, with clear and unobstructed faces, and captured under bright and sufficient lighting conditions. The four portraits are chosen to exhibit significant angular differences and expression variance, offering a range of prior perspectives. The video length is approximately 10 seconds, without shot cuts, typically sourced from interview footage due to its clarity of facial features and stable camera set.

\para
\noindent\textbf{Prompts.} To comprehensively evaluate the generative capabilities across diverse scenarios, we devise a systematic process of prompt construction for the IPT2V and IEPT2V tasks. We start with the construction for the IPT2V task, where all seven portrait-video elements are expected to change. We instruct LLM to generate 50 prompts in the predefined format comprising seven elements, with 25 for males and 25 for females. In the instructions, we emphasized the necessity for a wide variety of scenes and styles, such as indoor settings, landscapes, science fiction, fantasy, elegance, and grandeur. We also stress the need for an appropriate degree of change in characters' expressions and movements. Subsequently, we randomly remove descriptions of hairstyle, clothing, or accessory from the prompts to generate 50 differential prompts for the IEPT2V task. These prompts can be randomly combined with sets of visual references to generate up to 750 evaluation cases for each task. In the evaluation of our paper, we sample 50 sets of IPT2V cases and 50 sets of IEPT2V cases to compute metrics and draw comparisons, resulting in Tab.~\ref{tab} and Fig.~\ref{fig:main}. Note that when for the baselines, we transform the prompts into a format suitable for their text encoders by simplifying or paraphrasing them. Here, to briefly showcase the format of the generated prompts, three examples are provided below:
\begin{itemize}[leftmargin=*,labelindent=20pt] 
    \item Handheld shot, medium shot. The person has short, messy ash-blonde hair with bangs casually falling over his forehead. He is wearing a navy-and-white striped sailor shirt. His expression is filled with curiosity and excitement, eyes wide open. Slowly, he tilts his head back, following something in the sky with his gaze, until his face is almost pointing upward toward the heavens. Then, he gradually lowers his head and flashes an open, cheerful smile at the camera. The character's action takes place by the railing on a seaside promenade, with the sea breeze gently blowing through his hair. One hand shields his eyes from the sun while the other points to the sky. The video setting is a sunny coastal boulevard, with a backdrop of the deep blue ocean and white seagulls. The lighting is strong midday sunlight, creating a fresh, free-spirited, and energetic summer atmosphere.
    \item Static shot, medium shot. The person has long, straight black hair tied neatly into a ponytail, adorned with an intricate golden crown. He wears a lavish red silk robe embroidered with golden dragon patterns. His expression is solemn and majestic, with sharp, piercing eyes that exude authority without effort. Slowly, he turns his head from the left side toward the front, his gaze softening from its initial intensity into a compassionate gentleness. The character’s posture shows him seated regally on a high-backed chair, hands folded across his chest. As he moves his head, his shoulders shift slightly, every motion deliberate and steady. The video setting is the terrace of an ancient palace, with red pillars in the background and distant rolling green mountains. The lighting comes from bright natural daylight, bathing his dignified face and highlighting the gold embroidery on his robes, evoking a grand, majestic, and historically rich ambiance.
    \item Push-in shot, transitioning from medium to close-up. The person has a sleek black high ponytail. She wears a form-fitting navy-blue pilot uniform paired with a flight helmet, the visor pushed up onto her head. Her expression initially reflects post-mission relaxation, with a hint of fatigue in her eyes. Upon hearing instructions through her communicator, her gaze sharpens instantly. Slowly turning her head, she lifts the corner of her mouth into a confident smile, ready to embrace the next challenge. The character's action depicts her seated inside the cockpit of a mech or combat aircraft. The video setting is a high-tech hangar, with other mechs undergoing maintenance in the background. The lighting consists of complex instrument panel glows within the cockpit, combined with the bright operational lights of the hangar, creating layered illumination full of tense anticipation for battle.
\end{itemize}

\section{Analysis of Multiple References}
\begin{figure*}[t]
\centering
\includegraphics[width=0.8\linewidth]{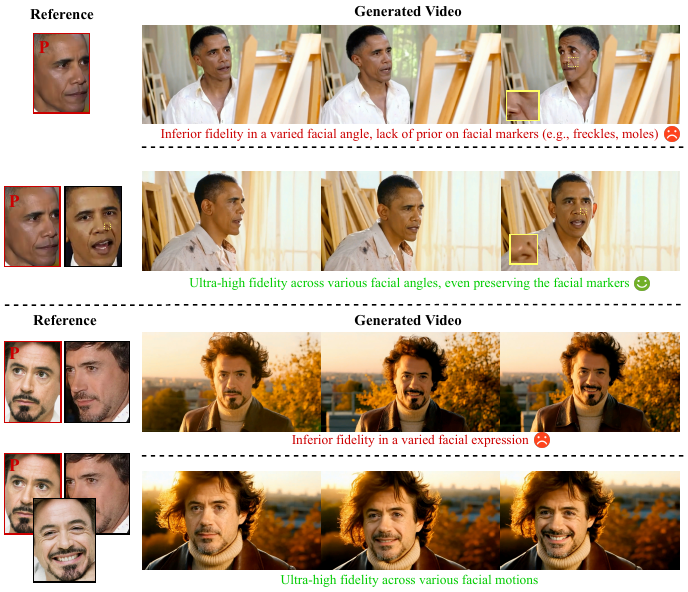}
\caption{Illustration of ablating the primary reference\&differential prompt design.}
\label{app:fig:multirefer}
\end{figure*}

The core contribution and motivation of AnyID is the introduction of multiple visual references. To further analyze the benefits brought by multiple references, we carry out a comparison in Fig.~\ref{app:fig:multirefer}, from which we can draw the following conclusions: (1) Multiple references can provide prior information on static facial features of the target person, including facial markers such as freckles and moles. Taking the particular example of "Obama" in Fig.~\ref{app:fig:multirefer}, for instance, he has a mole on one side of his nose. Only a single-angle reference lacks prior knowledge on this, making it completely impossible for the model to generate high-fidelity videos. This is the ill-posedness we have been emphasizing. By comparison, multiple references provide a holistic prior for the person, enabling the model to reconstruct the person with the highest possible fidelity. (2) Multiple references can provide prior information on dynamic features such as muscle movements and behavioral habits. As shown in the "Downey Jr." example in Fig.~\ref{app:fig:multirefer}, when references in the same expression are provided, the dynamics of the generated character will deviate. If we provide an additional reference for the expected expression, the results will improve significantly. 

In summary, multiple references can provide richer prior information, thereby significantly enhancing the fidelity of the model's output. Based on this, we also recommend providing references with a variety of expressions and angles when using our AnyID, rather than references with minimal variation.

\begin{figure*}[t]
\centering
\includegraphics[width=0.67\linewidth]{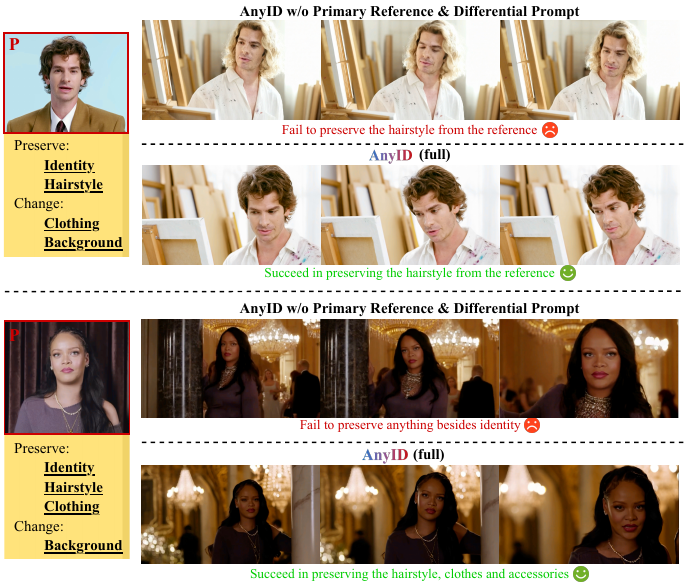}
\caption{Illustration of ablating the primary reference\&differential prompt design.}
\label{app:fig:ablation1}
\end{figure*}

\begin{figure*}[t]
\centering
\includegraphics[width=0.67\linewidth]{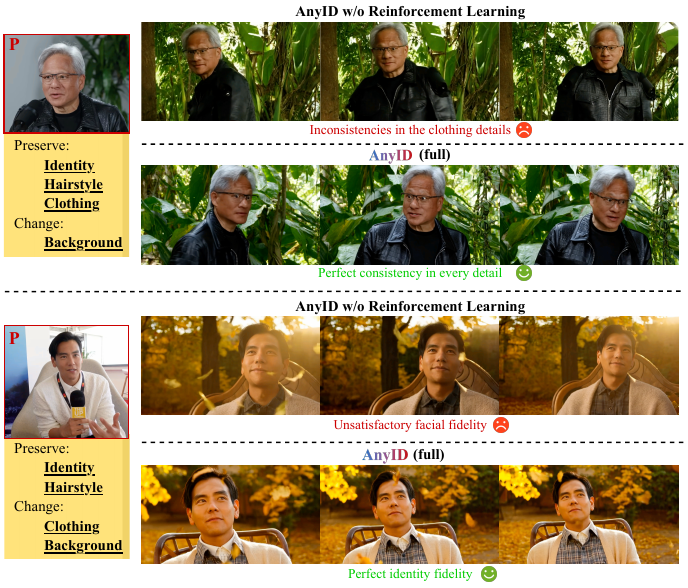}
\caption{Illustration of ablating the reinforcement learning.}
\label{app:fig:ablation2}
\end{figure*}

\section{Detailed Ablation Study}
\textbf{Primary Reference\&Differential Prompt Design.} Our AnyID features the design of a primary-reference generation and a differential prompt. The primary reference acts like a visual anchor, which provides element priors such as hairstyle and clothes. The differential prompt indicates the user-expected direction of modification, while the unmentioned aspects are expected to remain unchanged. Such design endows AnyID with the ability to follow the changes specified in the prompt and perfectly maintain the consistency of unmentioned elements. To prove this, we construct an ablating model of AnyID w/o primary reference\&differential prompt design and compare it with the full model in Fig.~\ref{app:fig:ablation1}. As shown, without such a design, the model's preservation of elements will be entirely controlled by the prompt, which is a weak constraint. As a result, the generations fail to preserve the hairstyle or the clothing from the reference. By comparison, the full model naturally anchors to the primary reference and succeeds in preserving the elements from it.

\para
\noindent\textbf{Reinforcement Learning.} AnyID introduces a human-centric reinforcement learning (RL) strategy to achieve the best identity fidelity and prompt controllability. To evaluate its effect, we illustrate the generations of AnyID w/o reinforcement learning (i.e., the model after supervised training) and the full model in Fig.~\ref{app:fig:ablation2}. As shown, the model without RL exhibits frequent flaws such as inconsistencies in details when preserving the elements or unsatisfactory facial fidelity when preserving the identity. By comparison, benefiting from the RL tuning based on human feedback, the full model stably demonstrates perfect identity fidelity, element consistency, and prompt controllability.

\begin{figure*}[t]
\centering
\includegraphics[width=0.744\linewidth]{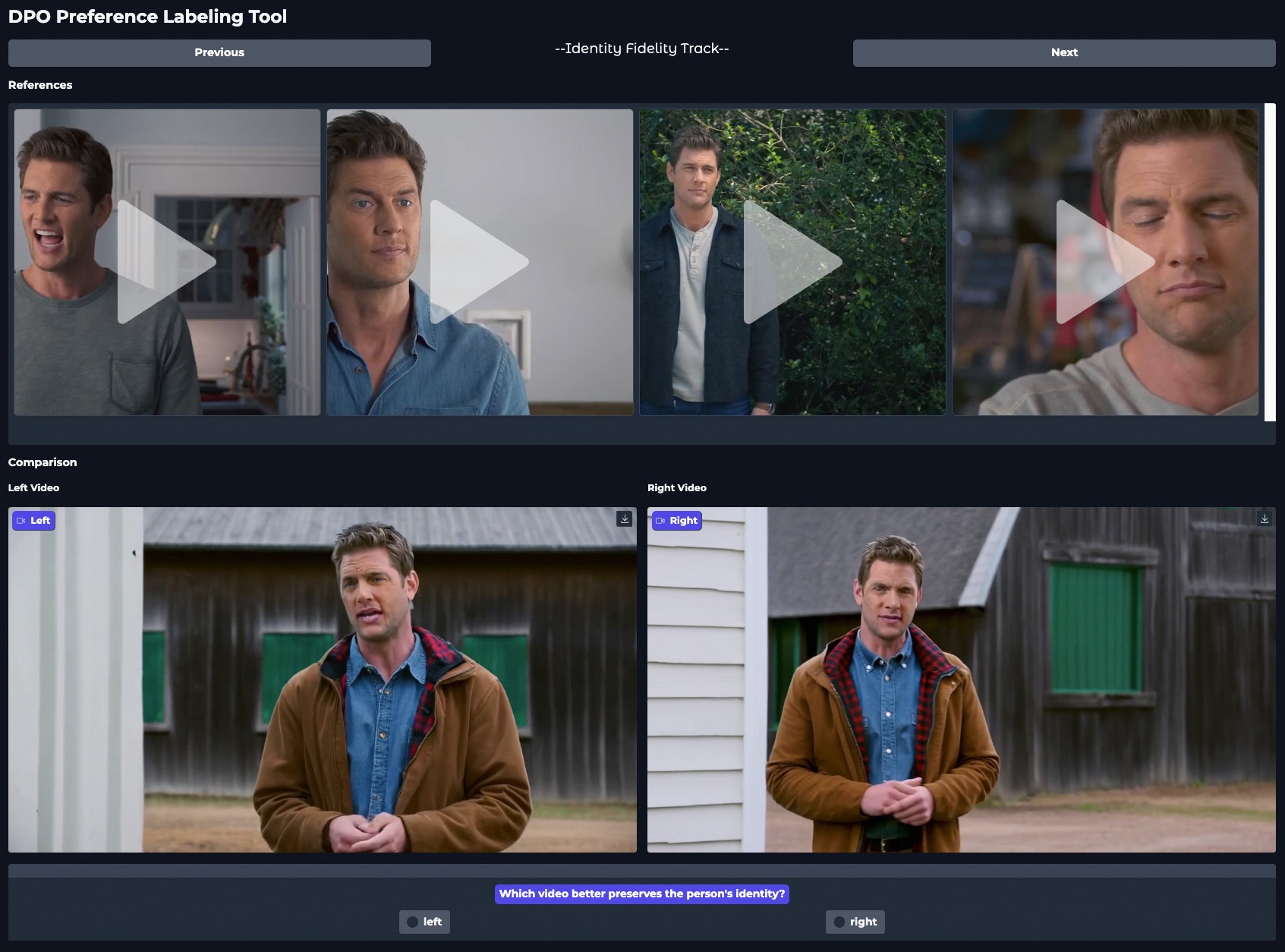}
\caption{A preference annotation example of the identity fidelity track.}
\label{app:fig:dpo1}
\end{figure*}
\begin{figure*}[t]
\centering
\includegraphics[width=0.744\linewidth]{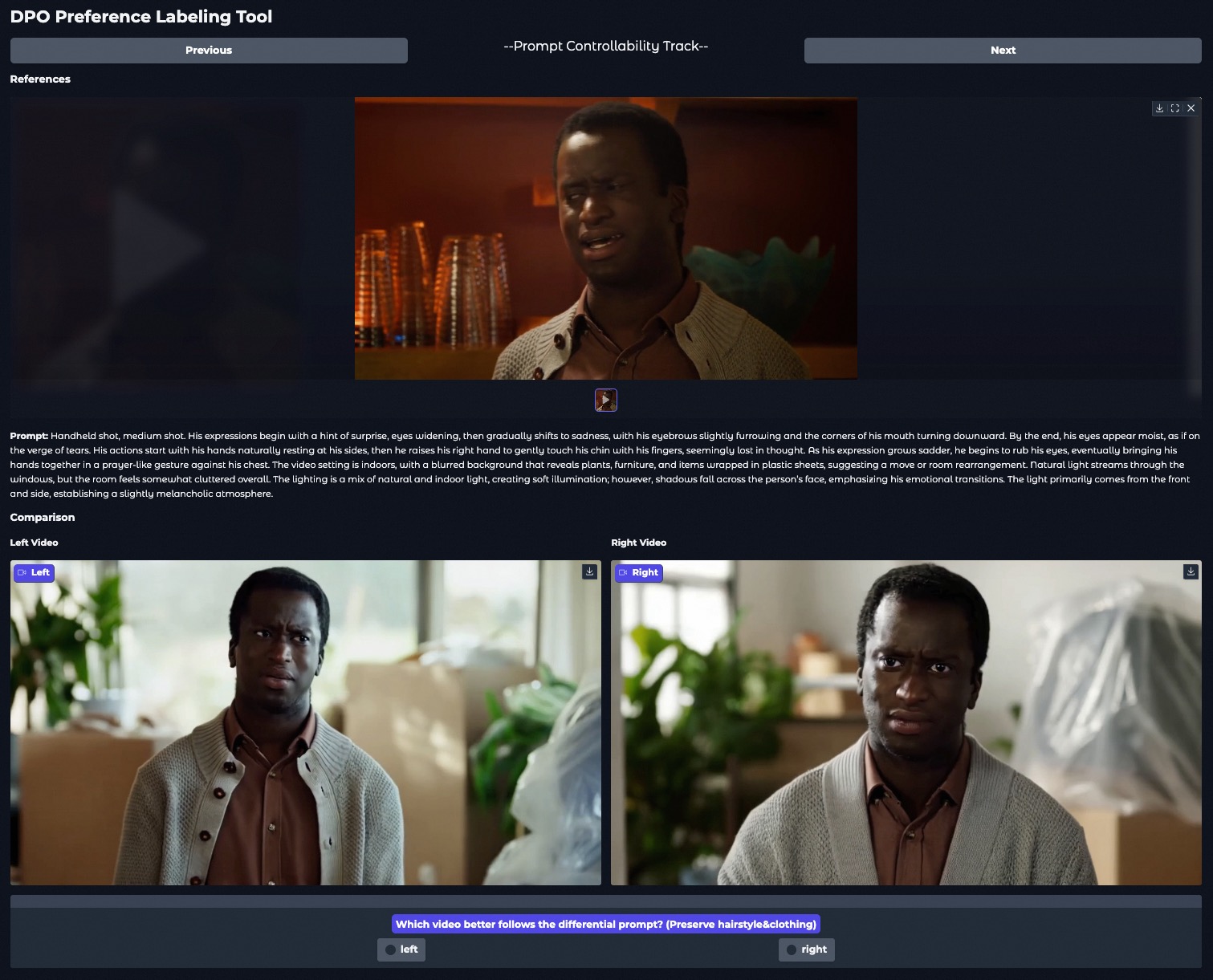}
\caption{A preference annotation example of the prompt controllability track.}
\label{app:fig:dpo2}
\end{figure*}

\section{Implementation of Preference Annotation}

The preference annotation process in our paper consists of two tracks: identity fidelity and prompt controllability. In practice, we build a frontend interface with two modes for these two tracks. Fig.~\ref{app:fig:dpo1} shows the identity fidelity mode, where all the clips from the source ID-group are provided to ensure that annotators have comprehensive prior information about the character's identity. The prompt is not provided to avoid distractions. Annotators need to choose the preferred video with higher fidelity from the two generation options. By comparison, Fig.~\ref{app:fig:dpo2} shows the prompt controllability mode, where only the primary reference and the differential prompt are used. Annotators need to choose the preferred video that better changes according to the prompt while preserving consistency in aspects not specified by the prompt.

\begin{figure*}[t]
\centering
\includegraphics[width=1\linewidth]{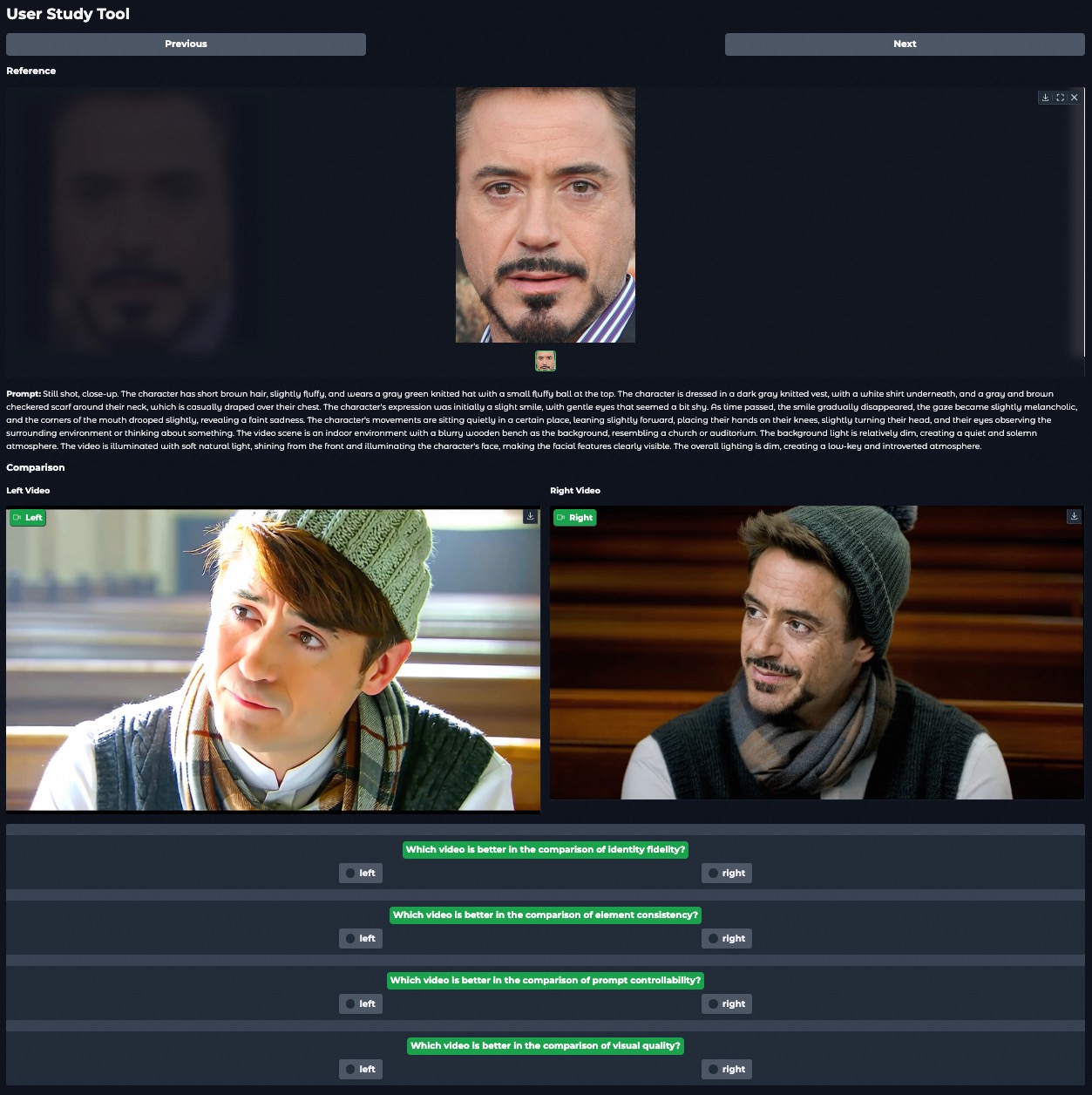}
\caption{An example of the user study.}
\label{app:fig:userstudy}
\end{figure*}

\clearpage
\section{Implementation of User Study}
To conduct the user study, we recruit volunteers and provide them with basic training based on the instructions below. We also build up a frontend interface shown in Fig.~\ref{app:fig:userstudy}. The user study is based on the evaluation set of celebrities. In each matchup, only the primary reference of the celebrity and the prompt are provided since the faces of celebrities are already widely recognized. Participants need to select the winner in four dimensions, under the guidance of the instructions.

\vspace{0.5cm}
\textbf{Instructions for Participants}

\vspace{0.5cm}
\hrule
\begin{quote}

\textbf{Thank you for participating in our study!}

We appreciate your willingness to contribute to our research. Below, you will find detailed instructions on how to complete the tasks involved in this study. Please read this information carefully before beginning.

\subsection*{1. Purpose of the Study}
The goal of this study is to evaluate different methods for generating identity-preserving portrait videos given a visual reference of a celebrity and a prompt. Your feedback will help us evaluate which method produces videos with the best quality. Your participation is highly valuable to us!

\subsection*{2. Task Overview}
You will be asked to compare two methods by evaluating their outputs in a series of one-on-one matchups. Each matchup will present results generated by our proposed method and a random baseline method (both anonymous). You will be given the visual reference along with the prompt. Note that the prompt only specifies the required changes, while aspects not mentioned in the prompt should remain consistent with the reference. Your task is to vote for your preferred option in each matchup based on the following criteria:
\begin{itemize}
    \item \textbf{Identity Fidelity}: Which method better preserves the dynamic identity of the celebrity (e.g., facial features, expressions, and overall appearance) across all frames and shots in the video?
    \item \textbf{Element Consistency}: Which method better preserves the consistency of the unmentioned elements, such as hairstyle or clothing? (Note that this item is optional; some matchups will have it while others will not.)
    \item \textbf{Prompt Controllability}: Which method more accurately changes the aspects specified in the given prompt (e.g., appearance, expressions, actions, and background) in the generated video?
    \item \textbf{Visual Quality}: Which method produces a video with higher visual quality, considering factors like sharpness, smoothness of motion, and absence of artifacts (e.g., blurriness or unnatural textures)?
\end{itemize}

\subsection*{3. How to Complete the Task}
\begin{itemize}
    \item You will complete \textbf{40 matchups} in total.
    \item For each matchup, you will see two outputs side by side.
    \item To cast your vote, click on the button corresponding to your preferred option.
    \item There are no right or wrong answers. Please choose the option that feels best to you.
\end{itemize}

\subsection*{4. Time Commitment}
The study should take approximately \textbf{20-30 minutes} to complete. You can work at your own pace, but we recommend completing the task in one sitting to ensure consistency.

\subsection*{5. Voluntary Participation}
Your participation in this study is entirely \textbf{voluntary}. You are free to withdraw at any time without penalty or explanation. If you decide to withdraw, your responses will not be included in the analysis.


\subsection*{6. Privacy and Data Use}
\begin{itemize}
    \item Your responses will be anonymized and used solely for research purposes.
    \item No personally identifiable information will be collected or stored.
\end{itemize}

\subsection*{7. Questions or Concerns}
If you have any questions about the study or encounter technical issues, please contact us. We are happy to assist you.

\subsection*{8. Consent}
By proceeding with the study, you confirm that:
\begin{itemize}
    \item You have read and understood the instructions.
    \item You agree to participate voluntarily.
    \item You understand that you can withdraw at any time without penalty.
\end{itemize}

Thank you again for your time and effort. Let’s get started!
    
\end{quote}
\hrule
\vspace{0.5cm}

\section{Societal Impacts and Safeguards}
The introduction of AnyID represents a significant leap forward in identity-preserving video generation, with far-reaching societal implications. By enabling ultra-fidelity synthesis of personalized characters from diverse and heterogeneous identity references—ranging from static portraits to dynamic video clips—AnyID empowers creators, educators, filmmakers, and developers to produce highly consistent, controllable, and expressive visual content with unprecedented ease. This capability lowers barriers to professional-grade video creation, fostering inclusivity in digital storytelling, virtual communication, gaming, and personalized education. Moreover, the fine-grained attribute-level control offered by our primary-referenced generation paradigm opens new avenues for creative expression, adaptive avatars, and immersive user experiences in metaverse and AR/VR applications.

Nevertheless, such powerful generative capabilities entail substantial societal risks that must be proactively addressed. The potential for malicious use—such as generating convincing deepfakes, non-consensual impersonations, or deceptive media—poses serious threats to individual privacy, public trust, and information integrity. Additionally, the automation of high-quality video production may disrupt traditional creative workflows, potentially displacing roles in video editing, animation, and performance capture, thereby necessitating workforce reskilling and ethical transitions in creative industries. To mitigate these concerns, AnyID incorporates robust safeguards inherited from the Wan foundation model suite, including built-in content moderation filters and output watermarking for provenance tracking. Besides, in alignment with responsible AI principles, we commit to full transparency: all code, models, and datasets will be publicly released under licenses that prohibit misuse and require attribution. By balancing innovation with ethical stewardship, we aim to ensure that AnyID serves as a force for creative empowerment while safeguarding against societal harm.

\end{document}